\documentclass[11pt]{article}

\usepackage[preprint]{acl}

\usepackage{times}
\usepackage{latexsym}

\usepackage[T1]{fontenc}

\usepackage[utf8]{inputenc}

\usepackage{microtype}

\usepackage{inconsolata}

\usepackage{graphicx}

\usepackage{amsmath}
\usepackage{amsfonts}
\usepackage{url}
\usepackage{booktabs}
\usepackage{graphicx}
\usepackage[table]{xcolor}
\usepackage{multirow}
\usepackage{amssymb}
\title{HyperGraphPro: Progress-Aware Reinforcement Learning \\ for Structure-Guided Hypergraph RAG}

\author{Jinyoung Park$^1$ \quad Sanghyeok Lee$^2$ \quad Omar Zia Khan$^3$\thanks{\hspace{0.5em} Work done while at Amazon.} \quad Hyunwoo J. Kim$^1$\thanks{\hspace{0.5em} Co-corresponding authors.} \quad Joo-Kyung Kim$^{4\dagger}$ \\
        $^1$KAIST, $^2$TwelveLabs,   $^3$Microsoft,   $^4$Amazon\\
        {\tt \{jinyoung.park, hyunwoojkim\}@kaist.ac.kr}\\ {\tt  sanghyeok.lee@twelvelabs.io}\\ {\tt  omarkhan@microsoft.com}\\ {\tt  jookyk@amazon.com}}
        
\definecolor{R1}{HTML}{56C596}       
\definecolor{R1-Searcher}{HTML}{4DA5E9}
\definecolor{Search-R1}{HTML}{3C88B6}  
\definecolor{Graph-R1}{HTML}{6A0DAD}
\definecolor{HyperGraphPro}{HTML}{AA2DAD}

\begin{document}
\maketitle
\begin{abstract}
Multi-hop question answering requires retrieval agents to construct a connected evidence path across reasoning turns rather than retrieve isolated relevant evidence for a single query.
However, in existing GraphRAG methods, retrieval and reinforcement learning are often optimized with mismatched signals: retrieval is driven by local semantic relevance, while outcome-level reinforcement learning assigns the reward by assessing the final answer without identifying which retrieval actions progress the reasoning path.
We introduce HyperGraphPro, a progress-aware hypergraph RAG framework that focuses on evidence-path progress as the shared objective for retrieval and policy optimization.
For retrieval, HyperGraphPro reranks candidate hyperedges using query-conditioned structural distinctiveness, combining entity-query relevance with a local-vs-global hypergraph incidence signal to favor entities that are informative in the current graph neighborhood.
For policy optimization, HyperGraphPro augments uniform trajectory-level reward assignment with stepwise progress-aware policy optimization, which assigns step-level rewards according to the coverage of answer-guided target hyperedges.
Experiments on multi-hop question answering benchmarks demonstrate that HyperGraphPro consistently improves reasoning accuracy over existing GraphRAG methods.
\end{abstract}

\section{Introduction}
Retrieval-augmented generation~(RAG)~\cite{lewis2020retrieval,jin2025search,song2025r1} has been widely used to improve the factuality of large language models~(LLMs) by providing output generation with external evidence.
However, conventional RAG approaches retrieve evidence as independent text chunks based on semantic similarity to the question.
It is often insufficient for multi-hop question answering, where the answer is not dependent on a single locally relevant text chunk but on a sequence of connected evidence paths.
For example, solving a compositional question may require iteratively identifying an intermediate entity, then following a relation to another entity to reach the final answer.
So, the challenge is not only retrieving relevant facts, but also constructing a coherent reasoning path with the connected evidence.

Graph retrieval-augmented generation~(GraphRAG) partially addresses this limitation by representing knowledge as entities and their higher-order relations rather than isolated knowledge chunks.
Recent agentic GraphRAG methods~\citep{luo2026graph} further allow language models to iteratively generate retrieval queries, observe graph-structured evidence, and update their reasoning state.

However, current agentic GraphRAG approaches still face a mismatch between local relevance and reasoning progress.
In retrieval, they largely rely on local contextual similarity, retrieving evidence that appears semantically relevant to the question, but may fail to capture the relational structure in the nature of graph structures.
Furthermore, the retrieved evidence across turns may remain disconnected.
This may cause the reasoning drift, where the agent provides plausible but unproductive evidence, accumulates redundant information, or misses bridging entities, that connect the question to the final answer.
In reinforcement learning, the objectives in GraphRAG rely only on sparse final-answer rewards without explaining which intermediate retrieval step advanced the reasoning path.
As a result, useful retrieval actions may receive little credit, while spurious actions in otherwise successful trajectories may be reinforced.

To address these limitations, we propose \textbf{HyperGraphPro}, which aligns both retrieval and policy optimization with graph-structured progress.
Our HyperGraphPro consists of two key components: structure-guided retrieval and progress-based stepwise policy optimization.
First, we introduce a \textbf{structure-guided hypergraph retrieval} mechanism, which reranks candidate
hyperedges by combining semantic relevance with query-conditioned structural distinctiveness.
This favors relational evidence whose query-relevant entities are concentrated within the query-induced hypergraph neighborhood.
Second, we present a \textbf{progress-based stepwise policy optimization} strategy, which provides dense supervision over intermediate reasoning steps by assessing the action of each step according to its contribution to reasoning progress, rather than relying solely on the correctness of the final answer.
Together, these components align both retrieval and policy learning with the compositional nature of graph-based reasoning.

We evaluate HyperGraphPro on 2WikiMultihopQA, HotpotQA, MuSiQue, and NQ using Qwen2.5-3B and Qwen2.5-7B backbones. HyperGraphPro consistently improves over the strongest evaluated baseline, Graph-R1, raising average F1 from 49.89 to 59.11 with Qwen2.5-3B and from 55.94 to 63.64 with Qwen2.5-7B.
The improvements are largest on multi-hop benchmarks, where the agent must connect evidence across retrieval steps. 
Our ablation studies further show that structure-guided retrieval and progress-aware policy optimization each improve performance and are complementary when combined.
Our contributions are threefold:
\begin{itemize}
    \item We propose a \textbf{query-conditioned structural distinctiveness score} for knowledge reranking, combining entity-query relevance with local-vs-global hypergraph incidence to prioritize informative entities in the current graph neighborhood.
    \item We introduce \textbf{progress-aware stepwise policy optimization}, which combines a coverage-augmented trajectory reward with turn-specific advantages based on coverage gains over answer-guided target hyperpaths.
    \item We demonstrate that \textbf{HyperGraphPro} achieves the best performance among the baselines including GraphRAG and RL-based agentic GraphRAG on multi-hop question answering benchmarks.
\end{itemize}
\section{Related Work}
\subsection{Graph Retrieval{\textendash}Augmented Generation}
Retrieval-Augmented Generation (RAG)~\citep{lewis2020retrieval,jiang2023active, trivedi2023interleaving} has been introduced to address hallucinated or factually incorrect responses generated by large language models.
GraphRAG~\citep{edge2024local} has extended the RAG paradigm by leveraging knowledge graphs to provide richer context and improve complex reasoning.
Building on GraphRAG, 
Hyper-RAG~\citep{feng2025hyper} and HypergraphRAG~\citep{luo2025hypergraphrag} harness the power of hypergraphs, enabling the capture of higher-order relationships beyond simple pairwise links.
In parallel, another line of work~\citep{chen2025pathrag,guo2025lightrag,gutierrez2025rag,wang2025causalrag,xu2025noderag}  explores retrieval mechanisms that emphasize computational efficiency without sacrificing reasoning accuracy.
More recently, Graph-R1~\citep{luo2026graph} further extends GraphRAG by leveraging reinforcement learning to enhance adaptability and reasoning performance.
Our HyperGraphPro presents structure-guided retriever that re-ranks candidate knowledge using query-conditioned structural distinctiveness. 

\subsection{Reinforcement Fine-Tuning for LLM}
Reinforcement learning (RL), such as RLHF~\citep{ouyang2022training}, has emerged as a key technique to boost LLM reasoning abilities, either by optimizing for outcome-based rewards or by using preference feedback. 
For example, DeepSeek-R1~\citep{guo2025deepseek} is trained with Group Relative Policy Optimization (GRPO), which evaluates multiple outputs from the model to estimate relative advantages, thereby eliminating the need for a separate value model for Generalized Advantage Estimation~\citep{schulman2015high}.
In addition, several works have explored agentic reinforcement learning frameworks~\citep{zheng2026deepeyes,feng2025group,zeng2025reinforcing}, enabling LLMs to autonomously decide when and how to invoke agents, thereby greatly improving their multi-turn interaction and tool-use capabilities.
Motivated by the success of reinforcement fine-tuning approaches, Search-R1~\citep{jin2025search} leverages multi-turn interleaved reasoning and search by dynamically retrieving external information, while R1-Searcher~\citep{song2025r1} employs a two-stage, outcome-based reinforcement learning scheme that enables LLMs to invoke search engines during reasoning.
\section{HyperGraphPro}

HyperGraphPro combines structure-guided hypergraph retrieval with progress-aware stepwise policy optimization.
The retriever uses query-conditioned entity distinctiveness to rerank relational evidence.
During training, the policy receives trajectory-level outcome feedback together with turn-specific signals derived from incremental target-hyperpath coverage.
We first introduce the agentic retrieval setting, then describe the retrieval module and the progress-aware training objective.
\subsection{Outcome-Level Agentic GraphRAG}
Reinforcement Learning~(RL)-based GraphRAG frameworks, such as Graph-R1~\citep{luo2026graph}, aim to learn an output reasoning trajectory $\boldsymbol{y}$ that leads to a final answer for an input question $\boldsymbol{x}$ by interleaving language reasoning with retrieval over the structured external knowledge graph $\mathcal{G}_K$.

\paragraph{Context similarity-based graph retrieval.}
GraphRAG leverages knowledge graphs~$\mathcal{G}_K$ to model the relational information inherent in the external knowledge.
In this work, we use a hypergraph as external knowledge representation since it expresses $n$-ary relations among more than two entities while conventional graphs model a connection between two entities.
Formally, a hypergraph is defined as $\mathcal{G}_K = \left(V, E\right)$, where $v \in V$ denotes entities and $E$ denotes a set of edges.
Generally, given the knowledge corpus ${K} = \left\{\boldsymbol{d}_1,\boldsymbol{d}_2 \dots, \boldsymbol{d}_N \right\}$, a (hyper)graph is extracted from knowledge documents $\boldsymbol{d}$ through an LLM:
\begin{equation}
    \mathcal{G}_K = \left(V, E \right), \text{where } \left\{\left(e_i, \mathcal{V}_{e_i} \right) \right\}_{i=1}^m \sim \pi_{\text{ext}}\left(\boldsymbol{d}\right)
\end{equation}
where $V, E$ are sets of nodes and hyperedges, respectively.
$\pi_{\text{ext}}$ is an LLM that is prompted to transform the knowledge segment $\boldsymbol{d}$ into a set of relation-entity pairs $\left\{\left(e_i, \mathcal{V}_{e_i} \right) \right\}$, where $e_i$ and $\mathcal{V}_{e_i}=\left\{v_1, \dots, v_n \right\}$ denote the hyperedge and its participating entities, respectively.
On the constructed knowledge graph, conventional methods~\citep{jin2025search} perform the retrieval based on their semantic similarity to the query representation.
It can be formulated as:
\begin{equation}
\begin{split}
&\mathcal{R}_{H}\left(\boldsymbol{q}_t \right) = \operatorname*{argmax}_{e_i \in E}^{k} \Big( \text{sim}(\phi(\boldsymbol{q}_t),\phi(e_i)) \Big), \\ &\mathcal{F}_{t}^{\text{ret}} = \bigcup_{e_j \in \mathcal{R}_{H}} \{(e_j, V_{e_j}) \mid  e_j \in E\},
\end{split}
\end{equation}
where $\phi(\boldsymbol{q}_t)$ denotes the query embeddings, and $\phi(e_i)$ represents the hyperedge embedding.
The top-$k$ retrieved set $\mathcal{F}_t^{\text{ret}}$ forms the final knowledge set.
Graph-R1 applies a hybrid hyperedge retrieval strategy where both entity and hyperedge similarity measurements are leveraged to retrieve hyperedges.

\paragraph{Group-Relative Policy Optimization.}
Group-Relative Policy Optimization~(GRPO)~\citep{shao2024deepseekmath, deepseekai2025deepseekv32} is one of the representative RL approaches with its strong performance and efficiency.
We apply GRPO to train the agent, which is formulated as:
\begin{equation}
\begin{split}
&\mathcal{J}_{\mathrm{GRPO}}(\theta)
= \mathbb{E}_{\left[\boldsymbol{x} \sim \mathcal{D}_X, \{\boldsymbol{y}^{(i)}\}_{i=1}^N \sim \pi_{\theta_{\text{old}}}(\cdot | \boldsymbol{x}; \mathcal{G}_K)\right]}
\\
&\Bigg[
\frac{1}{N}\sum_{i=1}^{N}
\frac{1}{L_i}\sum_{l=1}^{L_i}
\min\Big(
\rho_l^{(i)}\hat{A}^{(i)},
\\
&\qquad
\operatorname{clip}\!\left(
\rho_l^{(i)},\,1-\epsilon,\,1+\epsilon
\right)
\hat{A}^{(i)}
\Big)
\Bigg]\\
&-\beta \mathbb{D}_{\text{KL}}\left(
\pi_\theta \mid\mid \pi_{\text{ref}}
\right),
\end{split}
\end{equation}
where $\hat{r}_{t}^{(i)} = \frac{\pi_\theta\left( {y}^{(i)}_t \mid \boldsymbol{x}, \boldsymbol{y}_{<t}^{(i)}; \mathcal{G}_K\right)}{\pi_\text{old}\left(y_t^{(i)}\mid \boldsymbol{x}, \boldsymbol{y}_{<t}^{(i)}; \mathcal{G}_K\right)}$ denotes the likelihood ratio between current $\pi_\theta$ and the old policy model $\pi_{\text{old}}$. 
The advantage is calculated as $\hat{{A}}^{(i)} =  \frac{R\left(\boldsymbol{x}, \boldsymbol{y}^{(i)}\right)-\text{mean}\left(\left\{R\left(\boldsymbol{x},\boldsymbol{y}^{(j)} \right) \right\}_{j=1}^N \right)}{F\left(\left\{R\left(\boldsymbol{x},\boldsymbol{y}^{(j)} \right) \right\}_{j=1}^N \right)},$
where $F\left(\cdot \right)$ is the normalizer within the group~$\left\{R\left(\boldsymbol{x},\boldsymbol{y}^{(i)}
\right)\right\}_{i=1}^N$.
Most existing approaches apply outcome-supervised settings~\citep{guo2025deepseek}, which assign the same reward at every token in each output $\boldsymbol{y}$ based on the sequence-level reward of the output $\boldsymbol{y}$.

\subsection{Structure-Guided Retrieval for Evidence-Path Extension}
\label{sec:ret}

We introduce a structure-guided hypergraph retrieval module that ranks hyperedges using both semantic relevance and hypergraph structure.
The goal of retrieval in HyperGraphPro is not only to find hyperedges that match the current query, but to select hyperedges that extend the current evidence path.
A hyperedge is helpful when its entities are semantically compatible with the query and structurally informative in the local graph region induced by the query.
Conventional retrieval methods score each evidence unit independently, often relying on contextual similarity between the query and candidate passages.
Differently, our retriever asks whether a hyperedge contains entities that are not only semantically related to the query, but also structurally distinctive within the query-induced region of the hypergraph.

At reasoning step $t$, the agent generates a query $\mathbf{q}_t$ to the retriever.
We first identify a set of query entities, denoted by $\mathcal{V}(\mathbf{q}_t)$, and a candidate set of hyperedges, and then re-score the candidates by combining entity-level semantic relevance and structural distinctiveness.

\paragraph{Entity-query semantic relevance.}
We measure how well each entity $v \in V$ semantically aligns with the current query $\boldsymbol{q}_t$, which is computed as:
\begin{equation}
    s_t\left(v\right) = \mathrm{sim}\!\left(\phi(v), \phi \left(\mathcal{V}\left(\mathbf{q}_t\right)\right)\right), \qquad \forall v \in V,
\end{equation}
where $\phi(v)$ denotes the embedding of entity $v$, and
\begin{equation}
    \phi\left(\mathcal{V}\left(\mathbf{q}_t\right)\right) = \frac{1}{\left\lvert \mathcal{V}\left(\mathbf{q}_t \right)\right\rvert} \sum_{u \in \mathcal{V}\left(\mathbf{q}_t \right)} \phi \left(u \right)
\end{equation}
is the aggregated embedding of the entities mentioned in the query.
The score $s_t(v)$ captures the semantic compatibility between entity $v$ and the current required information.

\paragraph{Structure-guided entity distinctiveness.}
Semantic similarity alone is often insufficient in hypergraphs, because highly frequent entities may receive large similarity scores even when they provide limited discriminative value for multi-hop retrieval.
To address this issue, we introduce an entity distinctiveness term that reflects how distinctive an entity is within the query-induced hypergraph neighborhood by comparing its local participation frequency against its global participation frequency.

Assume that
\begin{equation}
    E(\mathbf{q}_t) = \{ e \in E \mid V_e \cap V(\mathbf{q}_t) \neq \emptyset \}
\end{equation}
denotes query-induced neighborhood, which is the set of hyperedges incident to at least one query entity, where $V_e$ denotes the entity set of hyperedge $e$.
For each entity $v \in V$, we define its query-conditioned distinctiveness of $v$ as
\begin{equation}
    D_t(v) = \log \left(
    1 +
    \frac{\left|\{ e \in E(\mathbf{q}_t) \mid v \in V_e \}\right|}
         {\left|\{ e \in E \mid v \in V_e \}\right|}
    \right).
\end{equation}
This score measures the fraction of an entity's incident
hyperedges that lie within the query-induced neighborhood.
In this sense, $D_t(v)$ acts as a query-conditioned structural prior that downweights overly common entities and amplifies entities that are more informative for the current retrieval context.

\paragraph{Hyperedge re-scoring.}
Given a candidate hyperedge $e \in E$, we first normalize the semantic relevance of each entity within the hyperedge:
\begin{equation}
    \tilde{s}_t(v,e) =
    \frac{s_t(v)}{\sum_{u \in V_e} s_t(u)}, \qquad v \in V_e.
\end{equation}
This normalization reflects the relative semantic contribution of entity $v$ within hyperedge $e$, rather than using its global relevance score in isolation.

We then combine semantic relevance and structural distinctiveness to obtain the entity contribution score
\begin{equation}
    r_t(v,e) = \tilde{s}_t(v,e)\cdot D_t(v),
\end{equation}
and define the overall score of hyperedge $e$ as
\begin{equation}
    R_t(e) = \sum_{v \in V_e} r_t(v,e) +\;  \mathrm{sim}\!\left(\phi(e), \phi(\mathbf{q}_t)\right).
\end{equation}
Therefore, $R_t(e)$ can be interpreted as the expected structural distinctiveness of the entities in hyperedge $e$, weighted by their semantic relevance to the current query.
A hyperedge receives a high score when its query-relevant entities are also distinctive within the query-induced hypergraph neighborhood.

Finally, we rank all candidate hyperedges by $R_t(e)$ and select the top-$k$ hyperedges:
\begin{equation}
    \mathcal{E}_t^{\mathrm{ret}} = \operatorname{TopK}_{e \in E}\, R_t(e).
\end{equation}
The retrieved fact set is then given by
\begin{equation}
    \mathcal{F}_t^{\mathrm{ret}}
    =
    \bigcup_{e \in \mathcal{E}_t^{\mathrm{ret}}}
    \{(e, V_e)\}.
\end{equation}
\subsection{Step Progress-Aware Policy Optimization for Evidence-Path Credit Assignment}
\label{sec:po}
RL-based retrieval-augmented agents~\citep{jin2025search,luo2026graph}
commonly optimize trajectory-level outcomes without explicitly
rewarding intermediate retrieval progress.
In outcome-based GRPO, a single trajectory-level advantage is
shared across reasoning turns.
For multi-hop reasoning over a knowledge hypergraph, however,
some retrievals recover missing relational evidence along a
question-to-answer path, whereas others add little to the
evidence already retrieved.
We therefore introduce \emph{hypergraph-aware stepwise policy
optimization}, which augments outcome supervision with the
incremental coverage of answer-guided target hyperpaths.
At each retrieval turn $t$, the agent generates a sub-query
$\boldsymbol{q}_t$ and retrieves hyperedges from the knowledge
hypergraph $\mathcal{G}_K$.

\paragraph{Hypergraph-aware progress rewards.}
We reward retrieval turns according to how much they expand the coverage of answer-guided hyperpaths.
Each hyperedge represents a relational fact involving multiple
entities and serves as the unit of evidence coverage.
This provides structure-aware supervision beyond conventional outcome-level rewards.
For each training question with target paths, let
$\{\mathcal{P}_b\}_{b=1}^{B}$ denote the answer-guided target
hyperpaths constructed in $\mathcal{G}_K$, where
$\mathcal{P}_b$ is the set of hyperedges constituting the
$b$-th path.
Specifically, the target hyperpaths are the shortest hyperpaths in $\mathcal{G}_K$, measured by the number of hyperedges, between the entities mentioned in the question and the gold answer entities, where consecutive hyperedges share at least one entity.
That is, $\mathcal{P}_b$ contains the hyperedges on a shortest path from the question to its answer, which form a minimal chain of relational facts that a successful retrieval trajectory should collect.
These target hyperpaths serve as structural supervision for evaluating the contribution of each retrieval turn.

For each target hyperedge $e$, let $c_t(e)$ denote the highest
rank-weighted matching score among all hyperedges retrieved
up to turn $t$, with $c_0(e)=0$.
When the entity of $e$ is included in one of the target hyperpaths, we score it with $(k+1-j)/k$, where $k$ is the number of retrieved hyperedges and $j$ is the retrieval rank of it.

We define the hypergraph-aware progress reward as the marginal gain in accumulated target-hyperpath coverage:
\begin{align}
    \Phi_t
    &=
    \frac{1}{B}
    \sum_{b=1}^{B}
    \frac{1}{|\mathcal{P}_b|}
    \sum_{e\in\mathcal{P}_b} c_t(e),
    \label{eq:turn_path_coverage}\\
    r_t^{\mathrm{sp}}
    &=
    \Phi_t-\Phi_{t-1}.
    \label{eq:step_progress_reward}
\end{align}
This reward captures incremental coverage of multi-entity
relational evidence along the target hyperpaths.
Retrievals that leave this coverage unchanged receive
zero progress reward.

\paragraph{Stepwise group-relative policy optimization.}
For each question $\boldsymbol{x}\sim\mathcal{D}_X$, we sample
a group of $N$ trajectories
$\mathcal{Y}=\{\boldsymbol{y}^{(i)}\}_{i=1}^{N}$
from
$\pi_{\theta_{\mathrm{old}}}
(\cdot\mid\boldsymbol{x};\mathcal{G}_K)$.
We obtain the sequence-level advantage $\hat{A}^{(i)}$
by standardizing the final rewards within this group.

Separately, we standardize the structure-progress rewards $r_t^{\text{sp}}$
across all retrieval turns in the same group to obtain
$z_t^{(i)}$,  which is computed as:
\begin{equation}
    z_t^{(i)} = \frac{r_t^{(i)}-\mu_t}{\sigma_t},
\end{equation}
where $\mu_t$ and $\sigma_t$ are computed over the corresponding
normalization set

The progress-modulated advantage is
\begin{equation}
    \widetilde{A}_t^{(i)}
    =
    \hat{A}^{(i)}
    +
    \lambda_{\mathrm{sp}} z_t^{(i)},
    \label{eq:step_modulated_advantage}
\end{equation}
where $\lambda_{\mathrm{sp}}$ controls the contribution of
turn-level progress.
For non-retrieval turns and questions without target hyperpaths,
we set $z_t^{(i)}=0$, recovering the sequence-level advantage.

Let $L_i$ denote the number of policy-generated tokens in
trajectory $i$, and let $t_i(l)$ map its $l$-th generated token
to the corresponding reasoning turn.
The token-level importance ratio is
$\rho_l^{(i)}
=
\pi_\theta(y_l^{(i)}\mid h_l^{(i)})
/
\pi_{\theta_{\mathrm{old}}}(y_l^{(i)}\mid h_l^{(i)})$,
where $h_l^{(i)}$ contains the question and the preceding
interaction history, including retrieved observations.
We maximize
\begin{equation}
\begin{split}
&\mathcal{J}_{\mathrm{GRPO}}(\theta)
= \mathbb{E}_{\left[\boldsymbol{x} \sim \mathcal{D}_X, \{\boldsymbol{y}^{(i)}\}_{i=1}^N \sim \pi_{\theta_{\text{old}}}(\cdot | \boldsymbol{x}; \mathcal{G}_K)\right]}
\\
&\Bigg[
\frac{1}{N}\sum_{i=1}^{N}
\frac{1}{L_i}\sum_{l=1}^{L_i}
\min\Big(
\rho_l^{(i)}\widetilde{A}_{t_i(l)}^{(i)},
\\
&\qquad
\operatorname{clip}\!\left(
\rho_l^{(i)},\,1-\epsilon,\,1+\epsilon
\right)
\widetilde{A}_{t_i(l)}^{(i)}
\Big)
\Bigg]\\
&-\beta \mathbb{D}_{\text{KL}}\left(
\pi_\theta \mid\mid \pi_{\text{ref}}
\right).
\end{split}
\end{equation}
All policy-generated tokens within a turn share its
progress-modulated advantage, while environment-inserted
tokens are excluded from the objective.
This provides turn-specific retrieval supervision while
retaining trajectory-level outcome feedback.

\begin{table*}[t!]
\centering
\resizebox{\textwidth}{!}{
\setlength{\tabcolsep}{1mm}{
\begin{tabular}{l cc cc cc |cc |cc | cc}
\toprule
\multirow{2.5}{*}{\textbf{Method}} & \multicolumn{2}{c}{\textbf{2Wiki.}} & \multicolumn{2}{c}{\textbf{HotpotQA}} & \multicolumn{2}{c}{\textbf{MuSiQue}} & \multicolumn{2}{|c|}{\textbf{Multihop Avg.}} & \multicolumn{2}{c|}{\textbf{NQ}} & \multicolumn{2}{c}{\textbf{All Avg.}} \\
\cmidrule(lr){2-3} \cmidrule(lr){4-5} \cmidrule(lr){6-7} \cmidrule(lr){8-9} \cmidrule(lr){10-11} \cmidrule(lr){12-13}
 & \textbf{EM} & \textbf{F1} & \textbf{EM} & \textbf{F1} & \textbf{EM} & \textbf{F1} & \textbf{EM} & \textbf{F1} & \textbf{EM} & \textbf{F1} & \textbf{EM} & \textbf{F1} \\
\midrule
\multicolumn{13}{l}{\textbf{\textit{Qwen2.5-3B-Instruct}}} \\
NaiveGeneration & 2.34 & 7.59 & 6.25 & 11.16 & 0.78 & 3.67 & 3.12 & 7.47 & 2.34 & 8.90 & 2.93 & 7.83 \\
StandardRAG & 3.91 & 12.52 & 7.03 & 15.41 & 0.00 & 2.92 & 3.65 & 10.28 & 0.00 & 10.69 & 2.74 & 10.39 \\
SFT & 7.03 & 12.40 & 10.94 & 16.48 & 1.56 & 5.04 & 6.51 & 11.31 & 3.12 & 11.23 & 5.66 & 11.29 \\
R1 & 20.31 & 28.45 & 20.31 & 25.33 & 3.12 & 8.07 & 14.58 & 20.62 & 11.72 & 21.51 & 13.87 & 20.84 \\
Search-R1 & 31.25 & 38.04 & 38.28 & 43.84 & 3.91 & 7.65 & 24.48 & 29.84 & 24.22 & 37.96 & 24.42 & 31.87 \\
R1-Searcher & 13.28 & 23.50 & 35.94 & 42.44 & 7.81 & 12.81 & 19.01 & 26.25 & 24.22 & 36.53 & 20.31 & 28.82 \\
Graph-R1 & 50.00 & 57.56 & 50.78 & 56.75 & 32.81 & 40.51 & 44.53 & 51.61 & 30.47 & 44.75 & 41.02 & 49.89 \\
\midrule
\rowcolor{HyperGraphPro!15} 
\textbf{HyperGraphPro (Ours)} & \textbf{63.28} & \textbf{69.76} & \textbf{62.50} & \textbf{67.59} & \textbf{42.19} & \textbf{52.26} & \textbf{55.99} & \textbf{63.20} & \textbf{32.81} & \textbf{46.84} & \textbf{50.20} & \textbf{59.11} \\
\midrule
\multicolumn{13}{l}{\textbf{\textit{Qwen2.5-7B-Instruct}}} \\
NaiveGeneration & 3.12 & 12.25 & 6.25 & 16.58 & 0.00 & 4.06 & 3.12 & 10.96 & 1.56 & 13.00 & 2.73 & 11.47 \\
StandardRAG & 7.81 & 12.75 & 10.16 & 21.10 & 0.78 & 4.53 & 6.25 & 12.79 & 1.56 & 15.97 & 5.08 & 13.59 \\
SFT & 11.72 & 20.28 & 19.53 & 27.59 & 5.47 & 10.02 & 12.24 & 19.30 & 5.12 & 19.02 & 10.46 & 19.23 \\
R1 & 25.00 & 30.99 & 31.25 & 37.05 & 7.03 & 14.53 & 21.09 & 27.52 & 16.41 & 28.45 & 19.92 & 27.76 \\
Search-R1 & 36.72 & 41.29 & 44.53 & 50.85 & 14.84 & 22.35 & 32.03 & 38.16 & 32.03 & 45.88 & 32.03 & 40.09 \\
R1-Searcher & 27.34 & 33.96 & 39.84 & 46.36 & 10.16 & 16.63 & 25.78 & 32.32 & 32.03 & 44.93 & 27.34 & 35.47 \\
Graph-R1 & 55.47 & 65.04 & 57.03 & 62.69 & 36.72 & 46.17 & 49.74 & 57.97 & {33.59} & 49.87 & 45.70 & 55.94 \\
\midrule
\rowcolor{HyperGraphPro!15}
\textbf{HyperGraphPro (Ours)} & \textbf{68.75} & \textbf{75.51} & \textbf{63.28} & \textbf{69.67} & \textbf{46.88} & \textbf{58.67} & \textbf{59.64} & \textbf{67.95} & \textbf{35.94} & \textbf{50.71} & \textbf{53.71} & \textbf{63.64} \\
\bottomrule
\end{tabular}%
}
}
\caption{
Experimental results on multihop question-answering benchmark datasets such as 2WikiMultiHopQa, HotpotQA, and MuSiQue, and a singlehop question-answering dataset such as NQ.
Multihop Avg. indicates the mean performance over multihop datasets.
All Avg. indicates the mean performance over all datasets.
}
\label{T2}
\end{table*}
\section{Experiments}
\subsection{Experimental Setup}
\noindent\textbf{Datasets.}

We evaluate our proposed methods on \textcolor{black}{three} multi-hop question answering benchmark datasets: 2WikiMultihopQA~(\textbf{2Wiki.})~\citep{ho2020constructing}, \textbf{HotPotQA}~\citep{yang2018hotpotqa}, \textbf{MuSiQue}~\citep{trivedi2022musique}.
We also adopt a large-scale single-hop QA dataset, Natural Questions (\textbf{NQ})~\citep{kwiatkowski2019natural} to explore the performance under the single-hop questions.
\begin{table*}[t]
    \centering
    \resizebox{\textwidth}{!}{
    \begin{tabular}{c|cccccc|cccccc}
        \toprule
           & \multicolumn{2}{c}{2Wiki.}&\multicolumn{2}{c}{MuSiQue}&\multicolumn{2}{c|}{Avg.}& \multicolumn{2}{c}{2Wiki.}&\multicolumn{2}{c}{MuSiQue}&\multicolumn{2}{c}{Avg.} \\
           \cmidrule(l{3pt}r{3pt}){2-3} \cmidrule(l{3pt}r{3pt}){4-5} \cmidrule(l{3pt}r{3pt}){6-7}\cmidrule(l{3pt}r{3pt}){8-9} \cmidrule(l{3pt}r{3pt}){10-11} \cmidrule(l{3pt}r{3pt}){12-13}
         Method   & EM & F1 &EM & F1 & EM & F1 & EM & F1 &EM & F1 & EM & F1  \\
            \midrule
        &\multicolumn{6}{l|}{\textbf{\textit{Qwen3-0.6B}}} & \multicolumn{6}{l}{\textbf{\textit{Qwen3-1.7B}}} \\
           Search-R1  & 33.59 & 37.92 & 9.38 & 15.55 & 21.49 & 26.74 & 31.25 & 36.79 & 10.94 & 17.23 & 21.10 & 27.01 \\
           Graph-R1   & 35.94 & 43.66 & 15.62 & 22.21 & 25.78 & 32.94 & 49.22 & 57.49 & 23.44 & 32.88 & 36.33 & 45.19 \\
            \rowcolor{HyperGraphPro!15}
          \textbf{HyperGraphPro (Ours)}  
          & \textbf{43.75} & \textbf{51.43} & \textbf{26.56} & \textbf{33.91} & \textbf{35.16} & \textbf{42.67} & \textbf{54.69} & \textbf{62.50} & \textbf{32.81} & \textbf{41.91} & \textbf{43.75} & \textbf{52.21} \\
        \bottomrule
    \end{tabular}
    }
    \caption{Experimental results using Qwen3-0.6B and Qwen3-1.7B. Avg. denotes the mean over 2Wiki. and MuSiQue.}
    \label{tab:qwen3}
\end{table*}
\begin{table}[t]
    \centering
    \resizebox{\columnwidth}{!}{
    \begin{tabular}{cccccccc}
        \toprule
           & & \multicolumn{2}{c}{2Wiki.}&\multicolumn{2}{c}{MuSiQue} \\
           \cmidrule(l{3pt}r{3pt}){3-4} \cmidrule(l{3pt}r{3pt}){5-6}
            Str. guided ret. &  Progress-aware PO & EM & F1 &EM & F1 \\
            \midrule
           &   & 50.00 &57.56  & 32.81& 40.51 \\
           \checkmark &   &{59.38} &66.04  & 39.06 & 49.91\\
            & \checkmark  &{57.81} &65.88  & 37.50& 46.22 \\
            \rowcolor{HyperGraphPro!15}
          \checkmark & \checkmark  
          &\textbf{63.28} &\textbf{69.76} & \textbf{42.19} &\textbf{52.26} \\
          
        \bottomrule
    \end{tabular}
    }
\caption{Ablation studies of our HyperGraphPro on multi-hop question answering datasets with Qwen-2.5-3B. (Str. guided ret.: Structure-guided hypergraph retrieval, Progress-aware PO: Progress-aware Step-level policy optimization).
    }
    \label{tab:abl2}
\end{table}

\begin{table}[t]
    \centering
    \resizebox{\columnwidth}{!}{
    \begin{tabular}{lcccccc}
        \toprule
           & \multicolumn{2}{c}{2Wiki.}&\multicolumn{2}{c}{MuSiQue} \\
           \cmidrule(l{3pt}r{3pt}){2-3} \cmidrule(l{3pt}r{3pt}){4-5} 
          Retrieval method    & EM & F1     & EM & F1  \\
         \midrule 
         No Retrieval      & 20.31& 28.45 &  3.12 & 8.07 \\
         Knowledge corpus~(Ctxt.) & 31.25 & 38.04 & 3.91 & 7.65  \\
         Hyperedge~(Ctxt.) & 50.78 & 58.13  & 32.03 & 40.31 \\
         Graph-R1~(Ctxt.)  & 50.00 & 57.56  &  32.81 & 40.51 \\
         \rowcolor{HyperGraphPro!15}
         \midrule
         \textbf{HyperGraphPro~(Ours)~(Ctxt.+Struct.)} &\textbf{59.38} &\textbf{66.04} & \textbf{39.06} &\textbf{49.91} \\
        \bottomrule
    \end{tabular}
    }
    \caption{Performance comparison according to the retrieval methods on multi-hop question answering datasets using Qwen-2.5-3B. Ctxt. indicates the natural language context-based relevance. Struct. indicates the graph structure-based relevance. For this experiment, we do not apply progress-aware PO for HyperGraphPro for the fair comparison.
    }
    \label{tab:ret}
\end{table}

\begin{table*}[!t]
    \centering
    \resizebox{0.9\textwidth}{!}{
    \begin{tabular}{lcccccccc}
        \toprule
           & \multicolumn{2}{c}{2Wiki.}&\multicolumn{2}{c}{HotpotQA}&\multicolumn{2}{c}{MuSiQue}&\multicolumn{2}{c}{NQ} \\
           \cmidrule(l{3pt}r{3pt}){2-3} \cmidrule(l{3pt}r{3pt}){4-5} \cmidrule(l{3pt}r{3pt}){6-7} \cmidrule(l{3pt}r{3pt}){8-9} 
          Method    & \# Turns & F1     & \# Turns & F1 & \# Turns & F1 & \# Turns & F1  \\
         \midrule 
         \textbf{\textit{Qwen2.5-3B-Instruct}}\\
         Graph-R1  & 2.63 & 57.56  & 3.65 & 56.75 & 3.79 & 40.51 & 1.21 & 44.75 \\
         \rowcolor{HyperGraphPro!15}
         \textbf{HyperGraphPro~(\textit{ours})} & \textbf{2.51} & \textbf{69.76} & \textbf{3.14} & \textbf{67.59} & \textbf{2.70} & \textbf{52.26} & \textbf{1.17} & \textbf{46.84} \\ 
         \midrule
         \textbf{\textit{Qwen2.5-7B-Instruct}}\\
         Graph-R1  & 2.72 & 65.04  & 3.38 & 62.69 & 3.92 & 46.17 & \textbf{1.13} & 49.87 \\
         \rowcolor{HyperGraphPro!15}
         \textbf{HyperGraphPro~(\textit{ours})} & \textbf{2.58} & \textbf{75.51} & \textbf{3.09} & \textbf{69.67} & \textbf{3.18} & \textbf{58.67} & \textbf{1.13} & \textbf{50.71} \\ 
        \bottomrule
    \end{tabular}
    }
    \caption{Full comparison on the number of turns between Graph-R1 and HyperGraphPro~(Ours).}
    \label{tab:eff_full}
\end{table*}

\noindent\textbf{Models.}
We conduct experiments using \texttt{Qwen-2.5-Instruct} (3B, and 7B)~\citep{qwen2025qwen25technicalreport} as the base LLMs following existing works~\citep{luo2026graph}.
For the retrieval, we employ \texttt{bge-large-en-v1.5}~\citep{xiao2024cpackpackedresourcesgeneral} following Graph-R1~\citep{luo2026graph}.
Additional implementation details are in the supplement.

\noindent\textbf{Baselines.}
To show the effectiveness of our HyperGraphPro over previous methods, we compare it with the baselines: \textbf{NaiveGeneration}, \textbf{StandardRAG}, \textbf{SFT}, \textbf{R1}~\citep{shao2024deepseekmath}, \textbf{Search-R1}~\citep{jin2025search}, \textbf{R1-Searcher}~\citep{song2025r1}, and \textbf{Graph-R1}~\citep{luo2026graph}.

\subsection{Experimental Results}
We evaluate our proposed methods using two variants of \texttt{Qwen2.5-Instruct}~(3B and 7B) in Table~\ref{T2}. 
HyperGraphPro consistently achieves the best average performance, showing remarkable improvements over both naive generation and retrieval-augmented baselines.
With \texttt{Qwen2.5-3B-Instruct}, our method reaches an average F1 of 59.11, compared to 49.89 from the strongest baseline (Graph-R1), a gain of 9.22 points.
Similarly, under \texttt{Qwen2.5-7B-Instruct}, our method achieves the highest overall performance with an average F1 of 63.64, improving upon the best baseline (Graph-R1, 55.94) by 7.70 points.
The improvement is more pronounced on multi-hop question-answering datasets than on NQ.
For \texttt{Qwen2.5-3B-Instruct}, the multihop average F1 gain over Graph-R1 is +11.59, compared with +2.09 on NQ.
For \texttt{Qwen2.5-7B-Instruct}, the multihop average F1 gain is +9.98, compared with +0.84 on NQ.
This result supports that progress-aware hypergraph retrieval and stepwise policy optimization are more beneficial when more connecting evidence is required across multiple reasoning turns.

\subsection{Additional Experimental Results}
\label{sec:additional}

\paragraph{Experimental results on Qwen3.}
To assess the effectiveness of our HyperGraphPro on other variants of LLMs, we conduct experiments using \texttt{Qwen3-0.6B} and \texttt{Qwen3-1.7B}~(Table~\ref{tab:qwen3}).
As shown in the table, HyperGraphPro consistently outperforms the baseline methods on both 2WikiMultiHopQA and MuSiQue for both Qwen3 variants.
On average, HyperGraphPro improves over the strongest baseline (Graph-R1) by +9.38 EM and +9.73 F1 with \texttt{Qwen3-0.6B}, and by +7.42 EM and +7.02 F1 with \texttt{Qwen3-1.7B}.
Notably, the performance gains are consistently observed for both EM and F1 scores, even with the substantially smaller 0.6B model.
These results demonstrate that the effectiveness of HyperGraphPro is maintained across different model scales, highlighting its robustness and general applicability to different LLM variants.

\paragraph{Ablation studies.}
To evaluate the effectiveness of the structure-guided hypergraph retrieval~(Sec.~\ref{sec:ret}) and progress-aware policy optimization~(Sec.~\ref{sec:po}), we conduct an ablation study on two multi-hop QA benchmarks in Table~\ref{tab:abl2}.
The results show that both components contribute to performance improvements on both datasets.
Specifically, applying progress-aware stepwise policy optimization alone improves F1 by +8.32 on 2Wiki. and +5.71 on MuSiQue, indicating better learning and enhanced reasoning ability.
Similarly, incorporating our structure-guided hypergraph retrieval framework alone yields consistent gains over the baseline (+8.48 and +9.40 F1 on 2Wiki. and MuSiQue, respectively), demonstrating the benefit of integrating hypergraph structural information during retrieval.
Then, combining both components, our full model achieves the best performance, with notable improvements in F1 scores of +12.20 on 2Wiki. and +11.75 on MuSiQue over the baseline.
These results confirm that both finer-grained policy updates and structure-aware retrieval are helpful for improving multi-hop question answering with large language models.

\begin{table*}[!t]
\tiny
\centering
    \resizebox{\textwidth}{!}{
    \begin{tabular}{p{2.8cm}|p{6.2cm}|p{5.4cm}}
        \toprule
        \textbf{Question \& Answer} &  \textbf{Graph-R1} & \textbf{HyperGraphPro~(Ours)} \\
        \midrule
          \textbf{Question: }Who is the chief minister in the state where Anjarle is located?
          \newline \textbf{Answer: }Devendra Fadnavis
            &
\textcolor{blue}{$\langle$think$\rangle$}
To find out who the chief minister is in the state where Anjarle is located, we first need to identify which state Anjarle is in.
\textcolor{blue}{$\langle$/think$\rangle$}\newline
\textcolor{orange}{$\langle$query$\rangle$}
which state is Anjarle located in?
\textcolor{orange}{$\langle$/query$\rangle$}\newline
\textcolor{green}{$\langle$knowledge$\rangle$}
Anjarle is a village in the Dapoli taluka of Ratnagiri district in the Maharashtra state of India. \textit{(rank 1)} ...
\textcolor{green}{$\langle$/knowledge$\rangle$}\newline
\textcolor{blue}{$\langle$think$\rangle$}
From the provided knowledge base, Anjarle is located in the Ratnagiri district of Maharashtra state. Now we need to find out who the chief minister is in the state of Maharashtra.
\textcolor{blue}{$\langle$/think$\rangle$}\newline
\textcolor{orange}{$\langle$query$\rangle$}
who is the chief minister of Maharashtra state?
\textcolor{orange}{$\langle$/query$\rangle$}\newline
\textcolor{green}{$\langle$knowledge$\rangle$}
Rank State Tax Revenues (INR Billions) 2010 - 2015 -- India. \textit{(rank 1)} ...
Prithviraj Chavan of the Congress party was the last Chief Minister of Maharashtra under the Congress/NCP alliance. \textit{(rank 7)} ...
\textcolor{green}{$\langle$/knowledge$\rangle$}\newline
\textcolor{blue}{$\langle$think$\rangle$}
According to the information provided, the chief minister of Maharashtra state is Prithviraj Chavan of the Congress party.
\textcolor{blue}{$\langle$/think$\rangle$}\newline
\textcolor{red}{$\langle$answer$\rangle$}
Prithviraj Chavan
\textcolor{red}{$\langle$/answer$\rangle$}
            &
\textcolor{blue}{$\langle$think$\rangle$}
To find the chief minister in the state where Anjarle is located, I first need to determine the state where Anjarle is situated. I can use a search query to find this information.
\textcolor{blue}{$\langle$/think$\rangle$}\newline
\textcolor{orange}{$\langle$query$\rangle$}
where is Anjarle located?
\textcolor{orange}{$\langle$/query$\rangle$}\newline
\textcolor{green}{$\langle$knowledge$\rangle$}
Anjarle is a village in the Dapoli taluka of Ratnagiri district in the Maharashtra state of India. \textit{(rank 1)} ...
\textcolor{green}{$\langle$/knowledge$\rangle$}\newline
\textcolor{blue}{$\langle$think$\rangle$}
From the knowledge base, I found that Anjarle is a village in the Dapoli taluka of Ratnagiri district in the Maharashtra state of India. Now, I need to find the chief minister of Maharashtra.
\textcolor{blue}{$\langle$/think$\rangle$}\newline
\textcolor{orange}{$\langle$query$\rangle$}
who is the chief minister of Maharashtra state?
\textcolor{orange}{$\langle$/query$\rangle$}\newline
\textcolor{green}{$\langle$knowledge$\rangle$}
Prithviraj Chavan of the Congress party was the last Chief Minister of Maharashtra under the Congress/NCP alliance. \textit{(rank 1)} ...
\textbf{Devendra Fadnavis} is the incumbent CM since 31 October 2014, the first from the BJP. \textit{(rank 8)}
With the exceptions of Manohar Joshi (Shiv Sena), Narayan Rane (Shiv Sena) and \textbf{Devendra Fadnavis} (Bharatiya Janata Party), all other CMs have been from the Congress or its breakaway parties. \textit{(rank 9)} ...
\textcolor{green}{$\langle$/knowledge$\rangle$}\newline
\textcolor{blue}{$\langle$think$\rangle$}
From the knowledge base, I found that the chief minister of Maharashtra is Devendra Fadnavis.
\textcolor{blue}{$\langle$/think$\rangle$}\newline
\textcolor{red}{$\langle$answer$\rangle$}
Devendra Fadnavis
\textcolor{red}{$\langle$/answer$\rangle$}\\
        \bottomrule
    \end{tabular}
    }
    \caption{Comparison on outputs generated by Graph-R1 and HyperGraphPro~(Ours) with Qwen2.5-3B.
    }
    \label{tab:qual}
    \vspace{-10pt}
\end{table*}

\paragraph{Retrieval method analysis.}
Table~\ref{tab:ret} compares different retrieval strategies on multi-hop QA datasets using \texttt{Qwen2.5-3B-Instruct}.
Our HyperGraphPro consistently achieves the best performance across all datasets, highlighting the effectiveness of integrating both contextual and structural relevance in retrieval.
Compared to the text-only retrieval baselines (Knowledge corpus and Hyperedge), HyperGraphPro yields substantial gains, demonstrating that leveraging graph structures enables more accurate and semantically coherent knowledge selection.
These results confirm that combining semantic and structural information is crucial for improving reasoning consistency in complex question answering.

\paragraph{Retrieval-turn analysis.}
We compare the average number of reasoning turns of HyperGraphPro with Graph-R1 using \texttt{Qwen-2.5-3B-Instruct} and \texttt{Qwen-2.5-7B-Instruct} in Table~\ref{tab:eff_full}.
The results show that HyperGraphPro achieves higher performance with fewer turns even though the number of turns is not explicitly used as a reward signal.
This demonstrates that our method improves reasoning efficiency while requiring fewer retrieval steps.
The reduction in turns is consistently observed across both the 3B and 7B models, particularly on MuSiQue, which requires complex multi-hop reasoning.
These results suggest that HyperGraphPro not only improves answer accuracy but also guides the model toward more focused and efficient retrieval trajectories.

\subsection{Qualitative Analysis}
We qualitatively compare Graph-R1 and HyperGraphPro using
\texttt{Qwen2.5-3B} on a two-hop question
(Table~\ref{tab:qual}).
Both agents correctly identify Maharashtra as the state
where Anjarle is located and then query its chief minister.
However, Graph-R1 bases its answer on a statement describing
Prithviraj Chavan as the last chief minister under the
Congress/NCP alliance, mistaking this historical reference
for evidence of incumbency.
In contrast, HyperGraphPro retrieves and uses evidence
explicitly identifying Devendra Fadnavis as the incumbent
in the retrieved context, even though the historical
statement about Chavan is ranked higher.
It therefore completes the reasoning chain from Anjarle
to Maharashtra and then to the chief minister, returning
the reference answer.
This example illustrates the importance of retrieving
and using evidence that supports the queried relation,
rather than relying on facts that merely mention the
relevant entities.
\section{Conclusion}

We have introduced HyperGraphPro, a step-aware reinforcement learning framework for graph-based retrieval and reasoning.
The core idea is to consider evidence-path progress in both retrieval and policy optimization.
While prior GraphRAG frameworks advanced agentic multi-step reasoning, they remain limited by structural unawareness and sparse reward feedback.
HyperGraphPro overcomes these issues through structure-aware retrieval and stepwise advantage modulation, enabling more coherent reasoning.
Experiments on knowledge-intensive QA benchmarks show that HyperGraphPro consistently improves answer accuracy on knowledge-intensive QA benchmarks.


\bibliography{main}

@inproceedings{ho2020constructing,
  title={Constructing a multi-hop QA dataset for comprehensive evaluation of reasoning steps},
  author={Ho, Xanh and Nguyen, Anh-Khoa Duong and Sugawara, Saku and Aizawa, Akiko},
  booktitle={COLING},
  pages= {6609--6625},
  year={2020},
}

@article{trivedi2022musique,
  title={{MuSiQue}: Multihop Questions via Single-hop Question Composition},
  author={Trivedi, Harsh and Balasubramanian, Niranjan and Khot, Tushar and Sabharwal, Ashish},
  journal={TACL},
  volume={10},
  pages={539--554},
  year={2022},
}

@inproceedings{trivedi2023interleaving,
  title={Interleaving retrieval with chain-of-thought reasoning for knowledge-intensive multi-step questions},
  author={Trivedi, Harsh and Balasubramanian, Niranjan and Khot, Tushar and Sabharwal, Ashish},
  booktitle={ACL},
  pages={10014--10037},
  year={2023},
}

@misc{qwen2025qwen25technicalreport,
      title={Qwen2.5 Technical Report}, 
      author={Qwen and : and An Yang and Baosong Yang and Beichen Zhang and Binyuan Hui and Bo Zheng and Bowen Yu and Chengyuan Li and Dayiheng Liu and Fei Huang and Haoran Wei and Huan Lin and Jian Yang and Jianhong Tu and Jianwei Zhang and Jianxin Yang and Jiaxi Yang and Jingren Zhou and Junyang Lin and Kai Dang and Keming Lu and Keqin Bao and Kexin Yang and Le Yu and Mei Li and Mingfeng Xue and Pei Zhang and Qin Zhu and Rui Men and Runji Lin and Tianhao Li and Tianyi Tang and Tingyu Xia and Xingzhang Ren and Xuancheng Ren and Yang Fan and Yang Su and Yichang Zhang and Yu Wan and Yuqiong Liu and Zeyu Cui and Zhenru Zhang and Zihan Qiu},
      year={2025},
      eprint={2412.15115},
      archivePrefix={arXiv},
      primaryClass={cs.CL}, 
}

@inproceedings{jiang2023active,
  title={Active retrieval augmented generation},
  author={Jiang, Zhengbao and Xu, Frank F and Gao, Luyu and Sun, Zhiqing and Liu, Qian and Dwivedi-Yu, Jane and Yang, Yiming and Callan, Jamie and Neubig, Graham},
  booktitle={EMNLP},
  year={2023}
}

@inproceedings{ouyang2022training,
  title={Training language models to follow instructions with human feedback},
  author={Ouyang, Long and Wu, Jeffrey and Jiang, Xu and Almeida, Diogo and Wainwright, Carroll and Mishkin, Pamela and Zhang, Chong and Agarwal, Sandhini and Slama, Katarina and Ray, Alex and others},
  booktitle={NeurIPS},
  volume={35},
  pages={27730--27744},
  year={2022}
}

@inproceedings{lewis2020retrieval,
  title={Retrieval-augmented generation for knowledge-intensive nlp tasks},
  author={Lewis, Patrick and Perez, Ethan and Piktus, Aleksandra and Petroni, Fabio and Karpukhin, Vladimir and Goyal, Naman and K{\"u}ttler, Heinrich and Lewis, Mike and Yih, Wen-tau and Rockt{\"a}schel, Tim and others},
  booktitle={NeurIPS},
  volume={33},
  pages={9459--9474},
  year={2020}
}

@article{xu2025noderag,
  title={NodeRAG: Structuring graph-based rag with heterogeneous nodes},
  author={Xu, Tianyang and Zheng, Haojie and Li, Chengze and Chen, Haoxiang and Liu, Yixin and Chen, Ruoxi and Sun, Lichao},
  journal={arXiv preprint arXiv:2504.11544},
  year={2025}
}

@article{schulman2015high,
  title={High-dimensional continuous control using generalized advantage estimation},
  author={Schulman, John and Moritz, Philipp and Levine, Sergey and Jordan, Michael and Abbeel, Pieter},
  journal={arXiv preprint arXiv:1506.02438},
  year={2015}
}

@inproceedings{zheng2026deepeyes,
  title={DeepEyes: Incentivizing" Thinking with Images" via Reinforcement Learning},
  author={Zheng, Ziwei and Yang, Michael and Hong, Jack and Zhao, Chenxiao and Xu, Guohai and Yang, Le and Shen, Chao and Yu, Xing},
  booktitle={ICLR},
  year={2026}
}

@inproceedings{luo2026graph,
  title={{Graph-R1}: Towards agentic graphrag framework via end-to-end reinforcement learning},
  author={Luo, Haoran and Chen, Guanting and Lin, Qika and Guo, Yikai and Xu, Fangzhi and Kuang, Zemin and Song, Meina and Wu, Xiaobao and Zhu, Yifan and Tuan, Luu Anh and others},
  booktitle={ICML},
  year={2026}
}

@inproceedings{jin2025search,
  title={{Search-R1}: Training llms to reason and leverage search engines with reinforcement learning},
  author={Jin, Bowen and Zeng, Hansi and Yue, Zhenrui and Yoon, Jinsung and Arik, Sercan and Wang, Dong and Zamani, Hamed and Han, Jiawei},
  booktitle={COLM},
  year={2025}
}

@inproceedings{song2025r1,
  title={Smart-Searcher: Incentivizing the Dynamic Knowledge Acquisition of LLMsvia Reinforcement Learning},
  author={Song, Huatong and Jiang, Jinhao and Min, Yingqian and Chen, Jie and Chen, Zhipeng and Zhao, Wayne Xin and Fang, Lei and Wen, Ji-Rong},
  booktitle={EMNLP Findings},
  year={2025}
}

@inproceedings{feng2025group,
  title={Group-in-group policy optimization for llm agent training},
  author={Feng, Lang and Xue, Zhenghai and Liu, Tingcong and An, Bo},
  booktitle={NeurIPS},
  year={2025}
}

@article{zeng2025reinforcing,
  title={Reinforcing Multi-Turn Reasoning in LLM Agents via Turn-Level Credit Assignment},
  author={Zeng, Siliang and Wei, Quan and Brown, William and Frunza, Oana and Nevmyvaka, Yuriy and Hong, Mingyi},
  journal={arXiv preprint arXiv:2505.11821},
  year={2025}
}

@inproceedings{gutierrez2025rag,
  title={{From RAG to Memory}: Non-parametric continual learning for large language models},
  author={Guti{\'e}rrez, Bernal Jim{\'e}nez and Shu, Yiheng and Qi, Weijian and Zhou, Sizhe and Su, Yu},
  booktitle={ICML},
  year={2025}
}

@inproceedings{wang2025causalrag,
  title={{CausalRAG}: Integrating causal graphs into retrieval-augmented generation},
  author={Wang, Nengbo and Han, Xiaotian and Singh, Jagdip and Ma, Jing and Chaudhary, Vipin},
  booktitle={ACL Findings},
  year={2025}
}

@inproceedings{chen2025pathrag,
  title={{PathRAG}: Pruning graph-based retrieval augmented generation with relational paths},
  author={Chen, Boyu and Guo, Zirui and Yang, Zidan and Chen, Yuluo and Chen, Junze and Liu, Zhenghao and Shi, Chuan and Yang, Cheng},
  booktitle={AAAI},
  year={2026}
}

@inproceedings{guo2025lightrag,
  title={{LightRAG}: Simple and fast retrieval-augmented generation},
  author={Guo, Zirui and Xia, Lianghao and Yu, Yanhua and Ao, Tu and Huang, Chao},
  booktitle={EMNLP},
  year={2025}
}

@article{edge2024local,
  title={From local to global: A graph rag approach to query-focused summarization},
  author={Edge, Darren and Trinh, Ha and Cheng, Newman and Bradley, Joshua and Chao, Alex and Mody, Apurva and Truitt, Steven and Metropolitansky, Dasha and Ness, Robert Osazuwa and Larson, Jonathan},
  journal={arXiv preprint arXiv:2404.16130},
  year={2024}
}

@article{feng2025hyper,
  title={{Hyper-RAG}: {Combating LLM Hallucinations using Hypergraph-Driven Retrieval-Augmented Generation}},
  author={Feng, Yifan and Hu, Hao and Hou, Xingliang and Liu, Shiquan and Ying, Shihui and Du, Shaoyi and Hu, Han and Gao, Yue},
  journal={Nature Communications},
  year={2026}
}

@inproceedings{luo2025hypergraphrag,
  title={{HyperGraphRAG}: Retrieval-Augmented Generation via Hypergraph-Structured Knowledge Representation},
  author={Luo, Haoran and Chen, Guanting and Zheng, Yandan and Wu, Xiaobao and Guo, Yikai and Lin, Qika and Feng, Yu and Kuang, Zemin and Song, Meina and Zhu, Yifan and others},
  booktitle={NeurIPS},
  year={2025}
}

@article{shao2024deepseekmath,
  title={Deepseekmath: Pushing the limits of mathematical reasoning in open language models},
  author={Shao, Zhihong and Wang, Peiyi and Zhu, Qihao and Xu, Runxin and Song, Junxiao and Bi, Xiao and Zhang, Haowei and Zhang, Mingchuan and Li, YK and Wu, Yang and others},
  journal={arXiv preprint arXiv:2402.03300},
  year={2024}
}

@inproceedings{yang2018hotpotqa,
  title={{HotpotQA}: A dataset for diverse, explainable multi-hop question answering},
  author={Yang, Zhilin and Qi, Peng and Zhang, Saizheng and Bengio, Yoshua and Cohen, William W and Salakhutdinov, Ruslan and Manning, Christopher D},
  booktitle={EMNLP},
  year={2018}
}

@article{kwiatkowski2019natural,
  title={Natural questions: a benchmark for question answering research},
  author={Kwiatkowski, Tom and Palomaki, Jennimaria and Redfield, Olivia and Collins, Michael and Parikh, Ankur and Alberti, Chris and Epstein, Danielle and Polosukhin, Illia and Devlin, Jacob and Lee, Kenton and others},
  journal={Transactions of the Association for Computational Linguistics},
  volume={7},
  pages={453--466},
  year={2019},
  publisher={MIT Press One Rogers Street, Cambridge, MA 02142-1209, USA journals-info~…}
}

@article{guo2025deepseek,
  title={Deepseek-r1: Incentivizing reasoning capability in llms via reinforcement learning},
  author={Daya Guo and Dejian Yang and Haowei Zhang and Junxiao Song and Ruoyu Zhang and Runxin Xu and Qihao Zhu and Shirong Ma and Peiyi Wang and Xiao Bi and Xiaokang Zhang and Xingkai Yu and Yu Wu and Z. F. Wu and Zhibin Gou and Zhihong Shao and Zhuoshu Li and Ziyi Gao and Aixin Liu and Bing Xue and Bingxuan Wang and Bochao Wu and Bei Feng and Chengda Lu and Chenggang Zhao and Chengqi Deng and Chenyu Zhang and Chong Ruan and Damai Dai and Deli Chen and Dongjie Ji and Erhang Li and Fangyun Lin and Fucong Dai and Fuli Luo and Guangbo Hao and Guanting Chen and Guowei Li and H. Zhang and Han Bao and Hanwei Xu and Haocheng Wang and Honghui Ding and Huajian Xin and Huazuo Gao and Hui Qu and Hui Li and Jianzhong Guo and Jiashi Li and Jiawei Wang and Jingchang Chen and Jingyang Yuan and Junjie Qiu and Junlong Li and J. L. Cai and Jiaqi Ni and Jian Liang and Jin Chen and Kai Dong and Kai Hu and Kaige Gao and Kang Guan and Kexin Huang and Kuai Yu and Lean Wang and Lecong Zhang and Liang Zhao and Litong Wang and Liyue Zhang and Lei Xu and Leyi Xia and Mingchuan Zhang and Minghua Zhang and Minghui Tang and Meng Li and Miaojun Wang and Mingming Li and Ning Tian and Panpan Huang and Peng Zhang and Qiancheng Wang and Qinyu Chen and Qiushi Du and Ruiqi Ge and Ruisong Zhang and Ruizhe Pan and Runji Wang and R. J. Chen and R. L. Jin and Ruyi Chen and Shanghao Lu and Shangyan Zhou and Shanhuang Chen and Shengfeng Ye and Shiyu Wang and Shuiping Yu and Shunfeng Zhou and Shuting Pan and S. S. Li and Shuang Zhou and Shaoqing Wu and Shengfeng Ye and Tao Yun and Tian Pei and Tianyu Sun and T. Wang and Wangding Zeng and Wanjia Zhao and Wen Liu and Wenfeng Liang and Wenjun Gao and Wenqin Yu and Wentao Zhang and W. L. Xiao and Wei An and Xiaodong Liu and Xiaohan Wang and Xiaokang Chen and Xiaotao Nie and Xin Cheng and Xin Liu and Xin Xie and Xingchao Liu and Xinyu Yang and Xinyuan Li and Xuecheng Su and Xuheng Lin and X. Q. Li and Xiangyue Jin and Xiaojin Shen and Xiaosha Chen and Xiaowen Sun and Xiaoxiang Wang and Xinnan Song and Xinyi Zhou and Xianzu Wang and Xinxia Shan and Y. K. Li and Y. Q. Wang and Y. X. Wei and Yang Zhang and Yanhong Xu and Yao Li and Yao Zhao and Yaofeng Sun and Yaohui Wang and Yi Yu and Yichao Zhang and Yifan Shi and Yiliang Xiong and Ying He and Yishi Piao and Yisong Wang and Yixuan Tan and Yiyang Ma and Yiyuan Liu and Yongqiang Guo and Yuan Ou and Yuduan Wang and Yue Gong and Yuheng Zou and Yujia He and Yunfan Xiong and Yuxiang Luo and Yuxiang You and Yuxuan Liu and Yuyang Zhou and Y. X. Zhu and Yanhong Xu and Yanping Huang and Yaohui Li and Yi Zheng and Yuchen Zhu and Yunxian Ma and Ying Tang and Yukun Zha and Yuting Yan and Z. Z. Ren and Zehui Ren and Zhangli Sha and Zhe Fu and Zhean Xu and Zhenda Xie and Zhengyan Zhang and Zhewen Hao and Zhicheng Ma and Zhigang Yan and Zhiyu Wu and Zihui Gu and Zijia Zhu and Zijun Liu and Zilin Li and Ziwei Xie and Ziyang Song and Zizheng Pan and Zhen Huang and Zhipeng Xu and Zhongyu Zhang and Zhen Zhang},
  journal={Nature},
  volume={645},
  pages={633--638},
  year={2025}
}

@misc{deepseekai2025deepseekv32,
      title={DeepSeek-V3.2: Pushing the Frontier of Open Large Language Models}, 
      author={DeepSeek-AI},
      year={2025},
}

@inproceedings{xiao2024cpackpackedresourcesgeneral,
      title={C-Pack: Packed Resources For General Chinese Embeddings}, 
      author={Shitao Xiao and Zheng Liu and Peitian Zhang and Niklas Muennighoff and Defu Lian and Jian-Yun Nie},
      year={2024},
      booktitle={SIGIR},
      url={https://arxiv.org/abs/2309.07597}
}

\end{document}